\documentclass[twoside]{article}

\usepackage[accepted]{aistats2021}
\usepackage[round]{natbib}

\usepackage{amsmath}
\usepackage{graphicx}
\usepackage{amssymb}
\usepackage{placeins}
\usepackage{threeparttable}
\usepackage{bm}
\usepackage{soul}
\usepackage{url}
\usepackage{color}
\usepackage{lipsum}
\usepackage{algorithm}
\usepackage[noend]{algpseudocode}

\usepackage{algorithm}
\usepackage{algpseudocode}
\algrenewcommand\algorithmicrequire{\textbf{Input:}}
\algrenewcommand\algorithmicensure{\textbf{Output:}}
\usepackage{color}
\usepackage{subcaption}

\begin{document}

%

%

\twocolumn[

\aistatstitle{Noise Contrastive Meta-Learning for Conditional Density Estimation using Kernel Mean Embeddings}

\aistatsauthor{ Jean-Fran\c cois Ton \And Lucian Chan \And  Yee Whye Teh \And Dino Sejdinovic }

\aistatsaddress{ University of Oxford } ]

\begin{abstract}
Current meta-learning approaches focus on learning functional representations of relationships between variables, i.e. estimating conditional expectations in regression. In many applications, however, the conditional distributions cannot be meaningfully summarized solely by expectation (due to e.g. multimodality). We introduce a novel technique for meta-learning conditional densities, which combines neural representation and noise contrastive estimation together with well-established literature in conditional mean embeddings into reproducing kernel Hilbert spaces. The method shows significant improvements over standard density estimation methods on synthetic and real-world data, by leveraging shared representations across multiple conditional density estimation tasks.
\end{abstract}

\section{Introduction}
The estimation of conditional densities $p(y|x)$ based on paired samples $\{(x_i,y_i)\}_{i=1}^n$ is a general and ubiquitous task when modelling relationships between random objects $x$ and $y$. While standard regression problems focus on estimating the conditional expectations $\mathbb{E}[y|x]$ of responses $y$ given the features $x$, many scenarios require a more expressive representation of the relationship between $x$ and $y$. In particular, the distribution of $y$ given $x$ may exhibit multimodality or heteroscedasticity.
A simple example of such a relation between $x$ and $y$ can be seen in the equation of a circle, as for any given $x\in[-1,1]$ there are two potential values $\pm \sqrt{1-x^2}$ the model could regress on. In this case, any standard regression model would fail to capture the true dependence between $y$ and $x$, because clearly $\mathbb E[y|x]=0$. 

Thus conditional density estimation requires a flexible nonparametric model of the conditional density and not just the expectation. Estimating conditional densities becomes even more challenging when the sample size is small. Hence, we approach this problem from a meta-learning perspective, where we are faced with a number of conditional density estimation tasks, allowing us to transfer information between them via a shared learned representation of both the responses $y$ and the features $x$.

Our contribution can be viewed as a development which parallels that of Neural Processes (NP) \citep{garnelo2018neural} and conditional Neural Processes (CNP) \citep{garnelo2018conditional} in the context of regression and functional relationships. Our proposed method is applicable to a much broader set of relationships between random objects, i.e. cases where for a given $x$ there are multiple potential responses $y$.
To that end, we will make use of the framework of conditional mean embeddings (CME) of distributions into reproducing kernel Hilbert spaces (RKHSs) \citep{muandet2017kernel, song2013kernel}, which we discuss in Section 2. 
%
%
%

In the RKHS literature, the feature maps that yield kernel mean embeddings that fully characterize probability distributions correspond to the notion of characteristic kernels \citep{Sriperumbudur2011} and are infinite-dimensional. However, such kernels can often be too simplistic for specific tasks (e.g. a simple Gaussian RBF kernel is characteristic) \citep{wilson2016deep,wenliang2018learning}. Moreover, even though they give a unique representation of a probability distribution and can be a useful tool to represent conditional distributions\footnote{In particular, CME $\mathbb E[\phi_y(y)|x]$ can be used to estimate conditional expectations $\mathbb E[h(y)|x]$ for a broad class of functions $h$, namely functions in the RKHS determined by the feature map $\phi_y$.}, they do not yield (conditional) density estimates, as they are merely points in the RKHS, and it is not clear how to adopt them for such tasks. 

To address this challenge, we propose a technique based on noise contrastive estimation (NCE) \citep{gutmann2012noise}, which treats the CMEs as dataset features in a binary classifier discriminating between the true and artificially generated samples of $(x_i,y_i)$ pairs. Although the CME estimation for fixed feature maps is well understood \citep{song2013kernel, muandet2017kernel}, we are concerned with the challenge of linking our CME estimates back to the conditional density estimation (CDE) task. This requires \emph{learning} the feature maps, $\phi_x, \phi_y$, that define the CME, which in turn determines the binary classifier for NCE.

In particular, we propose to use neural networks to learn appropriate feature maps $\phi_x$ and $\phi_y$ by adopting the meta-learning framework, i.e. by considering a number of (similar) conditional density estimation tasks simultaneously. Our meta-learning setting is concerned with CDE using small amounts of data (e.g. in the synthetic data setting each dataset only has 50 points). In this small data regime, standard CDE methods cannot work well, even on simple 1D problems, despite having asymptotic (large data) guarantees. Hence, in this paper, we mainly focus on low dimensional problems and discuss the higher dimensional problem in the conclusion.

Following the meta-learning principle that test and train conditions must match \cite{Vinyals2016}, we consider the case where multiple small datasets are available to train the system, such that it is able to estimate the conditional density well on a new unseen (also small) dataset. This is effectively a system for sharing  statistical strength across multiple CDE problems. In contrast to prior state-of-the-art work on meta-learning for regression, i.e. (attentive) Neural Process \citep{garnelo2018neural,kim2019attentive}, our proposed method is the first that can capture multimodal and complex conditional densities.

The proposed method is validated on synthetic and real-world data exhibiting multimodal properties, namely the NYC taxi data used in \citep{trippe2018conditional} to model the conditional densities of dropoff locations given the taxi tips, as well as  the Ramachandran plots from computational chemistry \citep{gravzulis2011crystallography}, which represent relationships between dihedral angles in molecular structures. We demonstrate significant improvement in terms of held-out loglikelihood over "the" standard conditional density estimation methods, including those based on meta-learning.





\begin{algorithm*}
\caption{MetaCDE Training}
\label{alg:his}
\begin{algorithmic}[1]
	\Require{$\mathcal{D}_{context} = \{(x^{cq}_1, y^{cq}_1) \dots (x^{cq}_{m_{cq}}, y^{cq}_{m_{cq}}) \}_{q=1}^l$ and $\mathcal{D}_{target} = \{(x^{tq}_1, y^{tq}_1) \dots (x^{tq}_{m_{tq}}, y^{tq}_{m_{tq}}) \}_{q=1}^l$}
    \Ensure{Optimized $\phi_x^{\theta}$, $\phi_y^{\theta}$, $b_\theta$, Note that we drop the $()^\theta$ for ease of notation}
	\Statex
	\For {$q=1, \dots, l$}
	\State Compute $\widehat{\mathcal{C}}_{Y|X}$ using $\mathcal{D}^q_{context}$ \Comment{Eq.(\ref{CME})}
	\State Generate $\kappa$ fake samples $\{y_{ij}^{f}\}_{i=1}^\kappa$ from the fake distribution $p_{f}(y)$ for each $y_j^{tq}$
	\State Compute $\widehat{\mu}_{Y|X=x^{tq}_j}$, $j=1 \dots m_{tq}$ \Comment{Eq.(\ref{CME_target})}
	\State Now construct the respective \textit{scoring functions} $s_{\theta}(x^{tq}_j, y^{tq}_{j})$ and $s_{\theta}(x^{tq}_j, y^{f}_{ij})$ using $\phi_y$ \Comment{Eq. (\ref{s_theta})}
	\EndFor
	\State Optimize the parameters $\theta$ of the NNs $\phi_x$, $\phi_y$, $b_\theta$ using $Adam$ jointly over all tasks \Comment{Eq.(\ref{eq:logisticloss})}
\end{algorithmic}
\end{algorithm*}

\begin{algorithm*}
\caption{MetaCDE Testing}
\label{alg:his}
\begin{algorithmic}[1]
	\Require{$\mathcal{D}^*_{context} = \{(x^{cq*}_1, y^{cq*}_1) \dots (x^{cq*}_{m_{cq*}}, y^{cq*}_{m_{cq*}}) \}_{q=1}^l$},  $\mathcal{D}^*_{target} = \{(x^{tq*}_1) \dots (x^{tq*}_{m_{tq*}}) \}_{q=1}^l$
	\Require{Trained $\phi_x$, $\phi_y$, $b_{\theta}$}, y \Comment{Evaluating the CDE at $y$}
	\Ensure{$p_{\theta}^q(y|x^{tq*})$, for  $q=1,...,l$} \Comment{The conditional density functions for each task.}
	\Statex
	\For {$q=1, \dots, l$}
	\State Compute the CMEO $\widehat{\mathcal{C}}_{Y|X}$ using Eq.(3) with $\mathcal{D}^{q*}_{context}$ and trained $\phi_x$, $\phi_y$
	\State Compute $s^q_{\theta}(x^{tq*}, y)$ using  Eq.(10) with $\widehat{\mathcal{C}}_{Y|X}$ and trained $\phi_x$, $\phi_y$
	\State Compute the conditional density $p_{\theta}^q(y|x^{tq*})$ using Eq.(4) i.e. with  $s^q_{\theta}(x^{tq*}, y)$ and $b_{\theta}$
	\EndFor
\end{algorithmic}
\end{algorithm*}

\section{Background}
We first introduce the notation used throughout this paper. Let $\mathcal{D} = \{(x_j, y_j) \}_{j=1}^n$ be the observed dataset, with $x_j\in\mathcal X$ being the input and $y_j\in\mathcal Y$ being the output. We denote the learned RKHS of inputs $X$ and responses $Y$ by $\mathcal{H}_X$ and $\mathcal{H}_Y$ respectively. Kernels of $\mathcal{H}_X$ and $\mathcal{H}_Y$ are denoted $k_x(\cdot,\cdot)$ and $k_y(\cdot,\cdot)$, and the corresponding feature maps are $\phi_x(\cdot)$ and $\phi_y(\cdot)$, i.e. $k_x(x_1,x_2) = \langle \phi_x(x_1),\phi_x(x_2)\rangle$ and similarly for $k_y$. 


\subsection{Conditional Mean Embeddings (CME)}
Kernel mean embeddings of distributions provide a powerful framework for representing and manipulating probability distributions \citep{song2013kernel, muandet2017kernel}. Formally, given sets $\mathcal{X}$ and $\mathcal{Y}$, with a distribution $P$ over the random variables $(X, Y)$ taking values in $\mathcal{X} \times \mathcal{Y}$, the conditional mean embedding (CME) of the conditional distribution of $Y|X=x$, assumed to have density $p(y|x)$, and is defined as:
\begin{align}
    \mu_{Y|X=x} := \mathbb{E}_{Y|X=x}[\phi_y(Y)] = \int_{\mathcal{Y}} \phi_y(y) p(y|x)dy.
    \label{defcme}
\end{align}
Intuitively, Eq.\ref{defcme} allows us to represent a probability distribution $p(y|x)$ in a function space (RKHS), by taking the expectation of features $\phi_y(y) \in \mathcal H_Y$ under $p(y|x)$.
Hence, for each value of the conditioning variable $x$, we obtain an element $\mu_{Y|X=x}$ of $\mathcal H_Y$. 

Following \citep{song2013kernel}, the CME can be associated with the operator $\mathcal{C}_{Y|X}:\mathcal H_X \to \mathcal H_Y$, also known as the conditional mean embedding operator (CMEO) s.t.
\begin{align}
\mu_{Y|X=x} = \mathcal{C}_{Y|X} \phi_x(x)
\label{cme}
\end{align}
where $\mathcal{C}_{Y|X}:= \mathcal{C}_{YX} \mathcal{C}_{XX}^{-1}$,  $\mathcal{C}_{YX}:=\mathbb{E}_{Y,X}[\phi_y(Y) \otimes \phi_x(X)]$ and $\mathcal{C}_{XX}:=\mathbb{E}_{X,X}[\phi_x(X) \otimes \phi_x(X)]$.

As a result, \citep{song2013kernel} have shown that the finite sample estimator of $\mathcal{C}_{Y|X}$ based on the dataset $\{(x_j, y_j) \}_{j=1}^n$ can be written as
\begin{align}
\widehat{\mathcal{C}}_{Y|X} = \Phi_y (K + \lambda I)^{-1} \Phi_x^T
\label{CME}
\end{align}
where $\Phi_y := (\phi_y(y_1), \dots, \phi_y(y_n))$ and $\Phi_x := (\phi_x(x_1), \dots, \phi_x(x_n))$ are the feature matrices, $K := \Phi_x^T\Phi_x$ is the kernel matrix with entries $K_{i,j}=k_x(x_i,x_j):=\langle \phi_x(x_i),\phi_x(x_j)\rangle$, and $\lambda>0$ is a regularization parameter. 


In fact, when using finite-dimensional feature maps $\phi_x$ and $\phi_y$, the conditional mean embedding operator is simply a solution to a vector-valued ridge regression problem (regressing $\phi_y(y)$ to $\phi_x(x)$) \citep{grunewalder2012conditional}, which allows computation scaling linearly in the number of observations $n$. The Woodbury matrix identity allows us to have computations of either order $\mathcal{O}(n^3)$ or $\mathcal{O}(d^3)+\mathcal{O}(d^2n)$, where $d$ is the dimension of the feature map $\phi_x$. In the meta-learning setting, the dataset size $n$ is small and hence the CME can be efficiently computed.

Now that we can compute the CME for a given new $x$, i.e. embed the conditional distribution $p(y|x)$ into a RKHS, the problem is how we are going to model the conditional density using the CME. In order to do that, we use ideas from noise contrastive estimation.

\subsection{Noise Contrastive Estimation (NCE)}

The seminal work on noise contrastive estimation \citep{gutmann2012noise} allows converting density estimation into binary classification, via learning to discriminate between noisy artificial data and real data.

To understand the methodology more concretely, let us first assume that the true underlying density of the data is $p(y|x)$ and that the distribution of the fake/artificial data is $p_{f}(y)$ (taken to be independent of $x$). Following \citep{gutmann2012noise} we formulate the classification problem, by sampling $\kappa$ times more fake examples than the real ones, which are all fed together with their labels (True/Fake) into the classifier. Hence, the data arises from $\frac{1}{\kappa+1}p(y|x) + \frac{\kappa}{\kappa+1}p_f(y)$ and the probability that, conditioned on a $x$, any given $y$ comes from the true distribution is 
$$P(\text{True}|y, x) = \frac{p(y|x)}{p(y|x) + \kappa p_f(y)}.$$

Since our goal is to learn the true density $p(y|x)$, we construct the probabilistic classifier, by imposing a parameterized form of $p(y|x)$ as $p_\theta(y|x)$.
In particular, we consider a generic density model given by
\begin{align}
    p_{\theta}(y|x) = \frac{\exp(s_{\theta}(x, y))}{\int \exp(s_{\theta}(x, y')) dy'} 
=\exp(s_{\theta}(x, y)+b_\theta(x)).
\label{cde_model}
\end{align}
for some functions $b_\theta : \mathcal X \to \mathbb R$ and $s_\theta : \mathcal X \times \mathcal Y \to \mathbb R$, the latter is referred to as the scoring function, following terminology in \citep{mnih2012fast}.
Hence the model for our classifier will be 
\begin{align}
   P_{\theta}(\text{True}|y, x) := \frac{p_\theta(y|x)}{p_\theta(y|x) + \kappa p_f(y)}.
   \label{classifier}
\end{align}
Note that the parameters of the classifier are the parameters of $s_{\theta}$ and $b_{\theta}$.
Assuming for the moment that the learned probabilistic classifier $P_{\theta}(\text{True}|y, x)$ attains Bayes optimality, we can deduce the point-wise evaluations of the true conditional density $p_{\theta}(y|x)$ directly from expression Eq.\ref{classifier} by simply rearranging for the density $p_{\theta}(y|x)$. We would also like to highlight some recent theoretical results with regards to NCE that have been developed in \citep{arora2019theoretical, gutmann2012noise}.  \cite[Theorem1]{gutmann2010noise} state that one only requires the support of the noise distribution to cover the true density support to recover the underlying density.


In section 3, we will merge the idea of NCE together with CME to estimate the conditional density in the meta-learning setting.  In particular, we will link our problem of computing the conditional density from the CME to NCE, by relating CME to $s_{\theta}(x, y)$ as defined in Eq.\ref{cde_model}.


\subsection{Meta-Learning} \label{subsection:metalearning}
Meta-learning is a growing area of which allows machine learning models to extract information from similar problems, and use this extracted (prior) information to solve new unseen problems quickly. Meta-learning and multi-task learning differ in the following way according to \citep{Finn2019}:
\begin{itemize}
    \item ``\textit{The meta-learning problem: Given data / experience on previous tasks, learn a \textbf{new task} more quickly and/or more proficiently}"
    \item ``\textit{The multi-task learning problem: Learn all of the tasks more quickly or more proficiently than learning them independently}"
\end{itemize}
In our case we perform meta-learning, by sharing statistical strength across multiple CDE problems. Standard CDE models usually require a lot of prior information to be useful in low data settings. In meta-learning however, this prior information is learned through the meta-learning phase, where the model is  presented with multiple small datasets during training such that given a new unseen dataset, the model is able to estimate the density well. 

Before we introduce our method, we would like to outline the meta-learning experimental setup that we will be using throughout the experiments in Section 5. Our setup is closely related to the one introduced in the papers on Neural Process \citep{garnelo2018neural, garnelo2018conditional}. 

Let us define the set of tasks $\mathcal{T}=\{T_1, \dots, T_l\}$ to be a set of $l$ conditional density estimation tasks, such that $T_q$ corresponds to the dataset $\mathcal D^q= \{(x_i^q, y_i^q)\}_{i=1}^{m_q}$, where $x_i^q\in \mathcal X$ and $y_i^q \in \mathcal Y$ share the same domains across the tasks. During training, we split the task $T_l$ into a ``context set'' and a ``target set''. The context set is used to summarize the task, whereas the target set is used to evaluate the summary of the context set. Hence, we use the target set to compute the loss and update the parameters of our model. We train our model jointly across multiple tasks and we can thus use this shared information to make better inference on similar tasks at testing time. During testing time, we will only be presented with $m^{context}_{q}$ samples (i.e. only a context set) and are then asked to do inferences on any given set of target points. 

\begin{figure*}[htp]
    \centering
    \includegraphics[width=.65\textwidth]{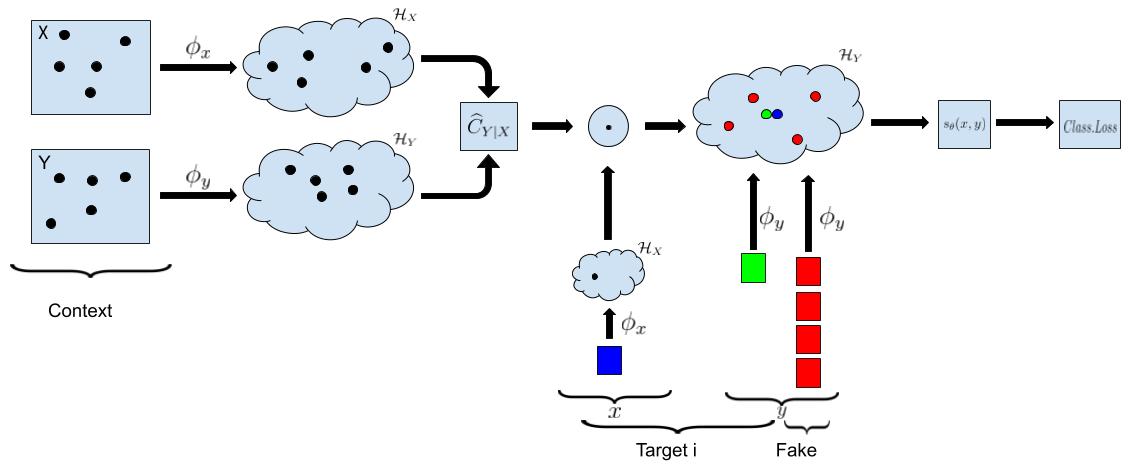}
    \caption{The context data is first passed through the feature maps to construct the CMEO ($\widehat{\mathcal{C}}_{Y|X}$), which is then used to project the new target $x$ (dark blue) to $\mathcal{H}_Y$. We then compare this projection to the \textit{True}(green) and \textit{Fake} $y$'s(red). Finally, we can compute the classification loss and back-propagate to update the parameters of the feature maps}
    \label{algo_fig}
\end{figure*}

\section{Methodology}

\subsection{Conditional Density Estimation}
As described in Section 2.2, the key ingredient of NCE is a classifier which can discriminate between the samples, i.e.  $\{y_i\}_{i=1}^n$, from the true density, in our case the conditional $p(y|x)$ , and fake samples, i.e. $\{y^f_i\}_{i=1}^{n\kappa}$, from the fake density $p_f(y)$.



Using the parameterization of Eq.\ref{cde_model} for our density model, the probabilistic classifier we will adopt is:
\begin{align}
  P_{\theta}(\text{True}|y, x) &=\frac{\exp(s_{\theta}(x, y)+b_\theta(x))}{\exp(s_{\theta}(x, y)+b_\theta(x)) +\kappa p_f(y)} \label{sigma_exp}\\
  &= \sigma\left(s_{\theta}(x, y)+b_\theta(x) - \log(\kappa p_f(y))\right) \nonumber
\end{align}
where $\sigma(t)=1/(1+e^{-t})$ is the logistic function. Learning the parameters $\theta$ of $s_{\theta}$ and $b_{\theta}$ through this classification allows us to learn the density in Eq.\ref{cde_model}, given that it defines the density. In the next section, we discuss the scoring function $s_{\theta}(x, y)$ and its relation to the feature maps $\phi_x$ and $\phi_y$.

Note that We only have approximations to the Bayes classifier, hence it will be useful following \citep{gutmann2012noise} to model the normalizing constant $b_{\theta}$ separately. We should also note that $b_\theta(x)=-\log \int \exp(s_{\theta}(x, y')) dy'$ from Eq.\ref{cde_model}. While the contribution $b_\theta(x)$ is directly determined by the choice of $s_\theta$, the calculation of $b_\theta$ is computationally intractable, and we therefore model $b_\theta(x)$ separately as it adds an extra independent component of $y$ and is suggested in the original noise contrastive paper \citep{gutmann2012noise}. We empirically note that learning $b_{\theta}$ through a different network helps training and keeps the learned density normalized (see Appendix for details). To make fair comparisons to other methods in terms of loglikelihood, we added an extra post-normalization step which is described in Section \ref{experiments}. For notation purposes, we collate all parameters of $s_{\theta}$ and $b_{\theta}$ into the parameter set $\theta$. 

\subsection{The choice of the scoring function $s_{\theta}(x, y)$}

We first map $x_i$ and $y_i$ using feature maps $\phi_x : \mathcal X \to \mathcal{H}_X$ and $\phi_y : \mathcal Y \to \mathcal{H}_Y$ respectively, which will be learned. We should note that we initially explored a fixed kernel choice with a characteristic kernel e.g. Gaussian RBF, but this resulted in poor empirical performance as the RBF was not flexible enough due to its restrictive stationarity constraints (see Appendix for experiments). Hence, in order to facilitate learning of these feature maps, we parametrized them with neural networks i.e. $\theta$. Note that in this case $\langle .,.\rangle_{\mathcal{H}_Y}$ is equivalent to a simple dot product between vectors.

Next, we compute the conditional mean embedding operator (CMEO) $\widehat{\mathcal{C}}_{Y|X}:\mathcal H_X \to \mathcal H_Y$ given in Eq.\ref{CME}. Using $\widehat{\mathcal{C}}_{Y|X}$, we can estimate the conditional mean embedding for any new $x^*$ as
\begin{align}
\widehat{\mu}_{Y|X=x^*} = \widehat{\mathcal{C}}_{Y|X} \phi_x(x^*).
\label{CME_target}
\end{align}
Note that $\widehat{\mu}_{Y|X=x^*} \in \mathcal{H}_Y$. Hence we can evaluate the function for any given new $y^*\in\mathcal Y$ by using the reproducing property of the RKHS, the linearity of the inner product and the definition in Eq.\ref{defcme},
\begin{align}
   \widehat{\mu}_{Y|X=x^*}(y^*) &=\langle \widehat{\mu}_{Y|X=x^*},  \phi_y(y^*)\rangle_{\mathcal{H}_Y} \\ 
    &=\int k_y(y^*,y)p(y|x^*)dy.
\end{align}
Intuitively, since a kernel $k_y$ expresses the similarity between its inputs,  we expect $k_y(y^*,y)$ to have high value when $y^*$ is drawn from the true distribution and low value when drawn from the fake distribution $p_f$, which thus affects the value of the CME accordingly.
This suggests the following form of the scoring function for our model: 
\begin{align}
s_{\theta}(x^*, y^*) = \langle \widehat{\mu}_{Y|X=x^*},  \phi_y(y^*)\rangle_{\mathcal{H}_Y}.
\label{s_theta}
\end{align}

Furthermore, we developed a purely neural version of our proposed method, namely MetaNN, in order to investigate the importance of the CMEO task representation. It differs from MetaCDE solely in the task representation. While MetaCDE uses kernel embeddings formalism to represent the task with CMEO calculation of the context points, MetaNN uses the DeepSets \citep{zaheer2017deep} approach, where the context pairs $(x_i, y_i)$ are simply concatenated into a vector, and then passed through a neural network. The outputs are then averaged to obtain the task embedding to which any new $x^*$ is concatenated to obtain the ``neural'' equivalent to CME. The same training procedure as MetaCDE follows.

We note that the concatenation of $x$ and $y$ encodes the joint distribution, rather than the conditional as in MetaCDE. While such task representation does preserve the relevant information, it is susceptible to changes in the marginal of $x$ across tasks and conditional representations are intuitively better suited for the task of conditional density estimation. The experiments demonstrate the value of combining the CME formalism with neural representations and we obtain significantly better results with MetaCDE compared to MetaNN. Additional details on MetaNN and MetaCDE can be found in the Appendix. 

Lastly, we also want to note that there are other choices for $s_{\theta}$ that could have been considered, i.e. $\epsilon$-KDE etc. However, the reason we opted for CME is firstly, its flexibility of parameterization using NN and secondly, its ability and theoretical background of capturing conditional densities in a RKHS.


\subsection{Training our proposed model}
For a given task $T_q$ corresponding to the dataset $\mathcal D^q= \{(x_i^q, y_i^q)_{i=1}^{m_q}\}$, we sample a set of fake responses $\{y_{i,j}^f\}_{i=1}^\kappa$ from $p_f$, associated to each $y_j$. In this case $y_{i,j}^f$ is the $i^{th}$ sample from the fake distribution for data point $y_j$. We can now train the classifier using the model given in Eq.\ref{sigma_exp} by maximizing conditional loglikelihood of the True/Fake labels, or equivalently, by minimizing the logistic loss:
\begin{align}\label{eq:logisticloss}
    \min_{\theta} \sum_{j=1}^{n}\bigg\{ \log \left(1+ \frac{\kappa p_f(y_j)}{\exp(s_{\theta}(x_j, y_{j})+b_{\theta}(x_j))}\right) + \\ \nonumber
   \sum_{i=1}^{\kappa}\log \left(1+ \frac{\exp(s_{\theta}(x_j, y^f_{i,j})+b_{\theta}(x_j))}{\kappa p_f(y^f_{i,j})}\right)\bigg\}. 
\end{align}

After parameters $\theta$ of the classifier have been learned, the conditional density estimates can be read off from Eq.\ref{cde_model}. Note that we need to be able to evaluate the fake density pointwise. Hence we consider a simple kernel density estimator (KDE) of $p(y)$ as our fake density, which we can easily evaluate pointwise. To sample from the this fake density $p_f$, we draw from the empirical distribution of all the $y$'s (context and target) and add Gaussian noise with standard deviation being the bandwidth of the KDE (we use a Gaussian KDE for simplicity). This simple approach yielded good results and hence we leave other methods, i.e. conditioning on $x$, for future work. 


We also note that Eq.\ref{eq:logisticloss} may be of an independent interest when learning feature maps for CMEs, i.e. where the goal is not necessarily density estimation, but other uses of CMEs discussed in \citep{song2013kernel}. Even though estimation of CME corresponds to regression in the feature space, it is inappropriate to use the squared error loss of the feature-mapped responses to learn the feature maps themselves. The reason being that the notion of distance in the loss is changing as the feature maps are learned and as a results they are not comparable across different feature maps. In fact, it would be optimal for the feature map $\phi_y$ to be constant, as the squared error would then be zero, without having learned anything about the relationship between $x$ and $y$.
\begin{table*}[!htp]
\centering
\resizebox{0.8\textwidth}{!}{%
\begin{tabular}{ll|lll}
                                    &         & Synthetic               & Chemistry                           & NYC                  \\ \hline
\multicolumn{1}{l|}{MetaCDE (Ours)} & Loglike & \textbf{197.84} $\pm$ \textbf{22.4}   & \textbf{-305.49} $\pm$ \textbf{46.9}             & \textbf{-1685.52}$\pm$ \textbf{608.35}  \\ \hline
\multicolumn{1}{l|}{MetaNN (Ours)}  & Loglike & 132.776$\pm$130.87    & -317.91$\pm$51.3               & -2276.55 $\pm$ 608.9 \\
\multicolumn{1}{l|}{}               & p-value & 4.781e-06               & 1e-03                           & 3.89e-10             \\ \hline
\multicolumn{1}{l|}{Neural Process} & Loglike & -81.11$\pm$18.5     & -426.75$\pm$ 47.3 & -3050.2 $\pm$ 822.8  \\
\multicolumn{1}{l|}{}               & p-value & \textless{}2.2e-16  & \textless 2.2e-16              & 3.89e-10             \\ \hline
\multicolumn{1}{l|}{DDE}            & Loglike & 162.98 $\pm$ 69.0   & -399.68 $\pm$ 41.3             & -2236.07 $\pm$ 565.9 \\
\multicolumn{1}{l|}{}               & p-value & 8.14e-07            & 1.65e-15                       & 3.89e-10             \\ \hline
\multicolumn{1}{l|}{KCEF}           & Loglike & -388.30 $\pm$ 703.1 & -724.40 $\pm$ 891.6            & -1695.89$\pm$435.4                   \\
\multicolumn{1}{l|}{}               & p-value & \textless 2.2e-16   & 9.72e-14                       & 0.025                  \\ \hline
\multicolumn{1}{l|}{LSCDE}          & Loglike & 44.95 $\pm$ 74.3    & -407.32 $\pm$ 80.1             & -2748.01 $\pm$ 549.2 \\
\multicolumn{1}{l|}{}               & p-value & \textless{}2.2e-16  & 2.57e-14                       &  3.89e-10             \\ \hline
\multicolumn{1}{l|}{e-KDE}          & Loglike & 116.31 $\pm$ 236.9  & -485.10 $\pm$ 303.4            & -2337.90 $\pm$ 501.1 \\
\multicolumn{1}{l|}{}               & p-value & 2.38e-07            & 2.94e-14                       &  4.13e-10            
\end{tabular}%
}
\caption{Due to the varying difficulties in tasks, the variance in loglikelihoods is high. Hence we added the p-values of a one-sided Wilcoxon test to show that MetaCDE is significantly outperforming competing methods in terms of loglikelihood} 
\label{table:results_table}
\end{table*}

\subsection{Meta-Learning of Conditional Densities}
The training procedure of our proposed method is described in Section \ref{subsection:metalearning}. Figure \ref{algo_fig} outlines one meta-training procedure for a given task, including the loss calculation. This step are repeated for every training task, as stated in Alg. 1. The procedures in testing time is stated in Alg. 2.

In this case, the CMEO, $\widehat{\mathcal{C}}_{Y|X}$ will be estimated using the context set, and the CMEs will be evaluated on the target set. The CMEO acts as the task embedding and the CME is used for the inferences on the target set. 

During training, for each target example, we sample $\kappa$ fake samples from $p_f(y)$ and represent them in $\mathcal{H}_Y$ using the feature map $\phi_y$. Hence we can construct $s_{\theta}$ so that Eq.\ref{sigma_exp} can be computed for each of these $\kappa+1$ samples ($1$ true and $\kappa$ fakes). By providing the labels (i.e. \textit{True}(from data)/\textit{Fake}(from noise distribution), we proceed by training the classifier. Thus we learn the parameters $\theta$ of neural networks $\phi_x^{\theta}$, $\phi_y^{\theta}$ and $b_\theta$ using the objective in Eq.\ref{eq:logisticloss} jointly over all tasks. 

The resulting feature maps hence generalize across tasks and can be readily applied to new, previously unseen datasets, where we are simply required to compute the scoring function $s_\theta(x,y)$ and normalization constant $b_{\theta}(x)$.

\subsection{Choice of the Fake Distribution in NCE}\label{sec:fakechoice}
The choice of the fake distribution plays a key role in the learning process here, especially because we are interested in conditional densities. In particular, if the fake density is significantly different from the marginal density $p(y)$, then our model could learn to distinguish between the fake and true samples of $y$ simply by constructing a ``good enough'' model of the marginal density $p(y)$ on a given task while completely ignoring the dependence on $x$ (this will then lead to feature maps that are constant in $x$). 

This becomes obvious if, say, the supports of the fake and the true marginal distribution are disjoint, where clearly no information about $x$ is needed to build a classifier -- i.e. the classification problem is ``too easy''. 

Thus, ideally we wish to draw fake samples from the true marginal $p(y)$ in a given task and hence, we propose to use a kernel density estimate (KDE) of $y$'s as our fake density in any given task.



In particular, a kernel density estimator of $p(y)$ is computed on all responses $y$ (context and target) during training. In our experiments we demonstrate that this choice ensures that the fake samples are sufficiently hard to distinguish from the true ones, requiring the model to learn meaningful feature maps which capture the dependence between $x$ and $y$ and are informative for the CDE task.



\section{Related Work}

Noise contrastive estimation for learning representations has been considered in the setting of Natural Language Processing (NLP). \cite{mnih2012fast} and \cite{ma2018noise} only focus on learning discrete distributions in the context of NLP using NCE. They achieved impressive speedups over other word embedding algorithms as they avoid computing the normalizing constant. 

More recently, \citep{oord2018representation} introduced a NCE method for representation learning, however, they focused on learning an expressive representation in the unsupervised setting, by optimizing a mutual information objective. Since they focused on representation learning, they did not require to evaluate the fake density point-wise. An alternative method that  used the idea of fake examples was \citep{zhang2018metagan}, which trained a GAN in order to use the resulting discriminator for few-shot classification. Their focus was on representation learning, rather than density estimation.

RKHSs in density models have been proposed, for example  \citep{dai2018kernel, arbel2017kernel} considered training kernel exponential family models, where the main bottleneck was the calculation of the normalizing constant. \citep{dai2018kernel} exploited the flexibility of kernel exponential families to learn conditional densities and avoided computing the normalizing constants by solving so-called nested Fenchel duals. \citep{arbel2017kernel} trained kernel exponential family models using score matching criteria, which allowed them to bypass normalizing constant computation. Their method however required computing and storing the first- and second order derivatives of the kernel function for each dimension and each sample, and thus required $\mathcal{O}(n^2d^2)$ memory and $\mathcal{O}(n^3d^3)$ time, for $n$ datapoints in $d$ dimensions.

CME-based sampling methods, such as kernel herding \citep{chen2012super, kanagawaphd}, have also been proposed. These methods are interested in producing representative samples using the CME as a reference. However, we use the CME as a feature to model the conditional density.

Another work that has recently used CME as task embeddings is meta-CGNN \citep{ton2020meta}. However, \cite{ton2020meta} are mainly interested in the causal direction detection.

All the above methods, but the last one, assume access to a large dataset to train on, as most of the theoretical guarantees only hold in the large data setting. We, however are interested in scenarios, where we are presented with only little data i.e. around 50 datapoints at test time in our synthetic dataset experiments. Hence, the work that come closest to ours is the Neural Process \citep{garnelo2018neural} as they also consider the meta-learning setting for density estimation. However, our method extends on their methodology in two major ways. First of all, we embed our task into an CMEO which takes into account the input covariates and responses in the context sets. The NP on the other hand does not consider this distinction, but rather concatenates them before pushing them through a neural network, thus loosing valuable information on their conditional relationship. Secondly, the NP objective is constrained to Gaussian output conditioned on the latent variable. In contrast, our method is able to deal with any modality of distributions as we consider a nonparametric model for the density.


\begin{figure*}
\centering
    \begin{subfigure}{\textwidth}
    \includegraphics[width=\textwidth]{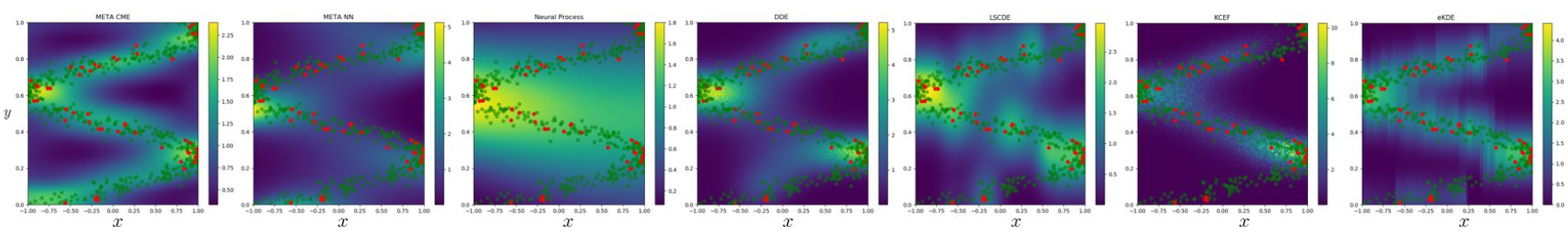}
    \end{subfigure}
    \begin{subfigure}{\textwidth}
    \includegraphics[width=\textwidth]{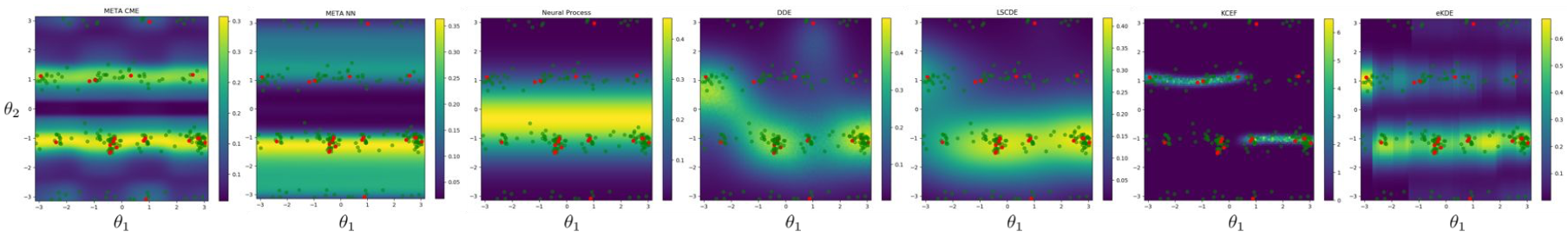}
    \end{subfigure}
    \caption{Density plots of Synth data (top) and Chem data (bottom). MetaCDE(ours), MetaNN(ours), NP, DDE, LSCDE, KCEF, $\epsilon$-KDE. Red: context points. Green: true density.}
    \label{fig:exp}
\end{figure*}

\section{Experiments}\label{experiments}
\subsection{Experiments on Synthetic Data}
To assess the ability of our method in learning the multimodality and heteroscedasticity in the response variable, we construct a synthetic dataset as follows: we sample $y_i \sim$ Uniform$(0,1)$ and set $x_i = \cos(a y_i + b)+\epsilon_i$, where $a$ and $b$ vary between tasks, with noise $\epsilon_i\sim \mathcal{N}(0,\sigma^2)$ (See Appendix for examples of tasks). This corresponds to a 90 degree rotated cosine curve with noise, resulting multimodality of $p(y|x)$. 


Figure \ref{fig:exp} shows the comparison between our method and a number of alternative conditional density estimations methods, including $\epsilon$-KDE (KDE applied to the $\epsilon$-neighbourhood of $x$), DDE \citep{dai2018kernel}, KCEF  \citep{arbel2017kernel}, and LSCDE \citep{sugiyama2010least}. We also include meta-learning algorithms such as the Neural Process \citep{garnelo2018neural} and a pure neural network version of our framework, MetaNN. 


For the meta-learning algorithms, we use $50$ context points and $80$ target points for each task. We also fix $\kappa=10$ as suggested in \citep{gutmann2012noise} for all our experiments. At testing time, we evaluate the method on $100$ new tasks with $50$ context points each. We simply pass the new context points to our model, which can evaluate the density with a simple forward pass as in Eq.\ref{cde_model}. The non meta-learning baselines are trained on each of the 100 datasets separately. We have included additional experiments with different dataset size, as well as details on the neural network architectures, hyper-parameters and experimental setup for our experiments in the Appendix. 

In Table \ref{table:results_table}, we report the mean loglikelihood over the 100 different datasets. Note that the NCE does not give us a perfectly normalized density. To ensure a fair comparison between methods, we post-normalize the densities for all our experiments, i.e. whenever we compute a conditional density, $p(y^*|X=x)$, we create a grid, $\{y_i\}_{i=1}^{100}$, equally spaced evaluations $p(y_i|X=x)$ over the range of the data, and normalize the density to 1 before computing the loglikelihood at $y^*$.  We should also note that the post-normalization  is minimal, see Appendix for more details.


The reason for the high variance in some methods stems from the varying difficulty of tasks. Hence, we also report the p-values of the one-sided signed Wilcoxon test. This allows us to confirm that our likelihood of MetaCDE is statistically significantly higher than all the other methods, including meta-learning methods such as NP and MetaNN.
\subsection{Experiments on NYC taxi data}
Next, we illustrate our algorithm on real world data such as the NYC taxi dataset from January 2016\footnote{Data from: \url{https://www1.nyc.gov/site/tlc/about/tlc-trip-record-data.page}}. We are interested in estimating conditional densities $p_{\text{pickup}}(\text{dropoff}|\text{tips})$. Hence, for each task, i.e. given the pick up location, our goal is to model the dropoff density conditionally on the tip amount. Different tasks correspond to different pickup locations and thus different conditional densities.

During training, we use 200 context and 300 target points. At testing time, we are given 200 datapoints of dropoff locations for an unseen pickup location and model the conditional density based on those.

In Appendix we added figures to show how the density evolves as the tip amount increases. In particular, we see that given a pickup location in Brooklyn, as the tip amount increases the trips are more likely to end in Manhattan. Note that this pickup location has not been seen during training. We illustrate additional unseen pickup locations in the Appendix as well as additional information on the experimental setup. We compute the loglikelihood on 50 unseen pickup location and perform a one-sided Wilcoxon test as in the above section (see Table \ref{table:results_table}).
\subsection{Experiments on Ramachandran plots for molecules}
Lastly, we apply our algorithm on a challenging chemistry dataset. Finding all energetically favourable conformations for flexible molecular structures in both bound and unbound state is one of the biggest challenges in computational chemistry \citep{chemreview} as the number of possibilities increases exponentially with the number of degrees of freedom in a molecule. The distribution of a pair of dihedral angles in molecules can be represented by \emph{Ramachandran plots} \citep{mardia2013}, and the knowledge of the correlated dihedral angles is limited by the library curated by the chemists. 
Here, we attempt to apply MetaCDE in order to learn richer relationships between dihedral angles, which can improve the current sampling scheme. 

Our experimental data was extracted from crystallography database \citep{gravzulis2011crystallography}. A task $T_q$ consists of a pair of dihedral angles defined in a molecular fragment $q$, which is composed of 2 smaller fragments, F1 and F2. In this case, we have $\mathcal D^q= \{(\theta_{1,i}^q, \theta_{2,i}^q)_{i=1}^{m_q}\}$, where $\theta_{1,i}^q$'s are the dihedral angle of F1 and $\theta_{2,i}^q$'s are the corresponding dihedral angles of F2. For more clarification, see Appendix. The multimodality arises from the molecular reflectional and rotational symmetry. Even though this dataset is a 1D problem is constitutes an important problem in chemistry and we show that conventional methods are unable to capture the true underlying density of the Ramachandran plots.

In our meta-learning setup, we use 20 context and 60 target points. At testing time we are given 20 datapoints. We again fix $\kappa=10$ as suggested in \citep{gutmann2012noise} and evaluate our method using loglikelihood on 100 examples of pairs of dihedral angles, which are unseen during training. We perform a one-sided signed Wilcoxon test to confirm that MetaCDE achieves a significantly higher held-out mean loglikelihood than other methods, as presented in Table \ref{table:results_table}. Figure \ref{fig:exp} also shows that our method is able to successfully capture the multimodality in the data (i.e. 2 parallel lines), whereas other methods fail on this task, by either modelling the mean or focusing only on one mode. For more patterns see Appendix.
\section{Conclusions and Future Work}
We introduced a novel method for conditional density estimation in a meta-learning setting. We applied our method to a variety of synthetic and real-world data, with strong performance on applications in computational chemistry and NYC taxi data. Owing to the meta-learning framework, the experiments shows the ability of our method in capturing the correct density structure even when presented with small sample sizes at testing time. In contrast to Neural Process \citep{garnelo2018neural}, we construct the task embedding using a conditional mean embedding operators, coupled with features maps learned using noise contrastive estimation.

In this work we only considered low-dimensional problems (1D and 2D) similar to  \cite{garnelo2018neural, garnelo2018conditional}, mainly because even in this setting, standard method still fail. We have demonstrated significant advantages of our method MetaCDE against the competition in important real-world applications such as the Ramachandran plots in chemistry. One weakness of the proposed method is that it would not scale well in the dimension of $y$ given the post-normalization/NCE needed. We leave this to future work to amend.


Lastly, an interesting avenue would be in modelling conditional distributions in the reinforcement learning setting. In particular, \citep{lyle2019comparative} have shown the benefits of using distributional perspective on reinforcement learning as opposed to only modelling expectations of returns received by the agents.

\section{Acknowledgments}
We would like to thanks Anthony Caterini, Qinyi Zhang, Emilien Dupont, David Rindt, Robert Hu, Leon Law, Jin Xu, Edwin Fong and Kaspar Martens for helpful discussions and feedback. JFT is supported by the EPSRC and MRC through the OxWaSP CDT programme (EP/L016710/1). YWT and DS are supported in part by Tencent AI Lab and DS is supported in part by the Alan Turing Institute (EP/N510129/1). YWT’s research leading to these results has received funding from the European Research Council under the European Union’s Seventh Framework Programme (FP7/2007-2013) ERC grant agreement no. 617071.

\clearpage

\twocolumn[]
\bibliography{ref}

\clearpage
\appendix
\section{Synthetic dataset setup and further experiments}
In order to measure the ability of our method metaCDE to learn multimodality and heteroscedasticity in the response variable, we construct the datasets as follows: we sample $y_i \sim$ Uniform$(0,1)$ and set $x_i = \cos(a y_i + b)+\epsilon_i$, where $a$ and $b$ vary between tasks, with noise $\epsilon_i\sim \mathcal{N}(0,\sigma^2)$.
Here we sample $a$ from $U(8, 12)$ and phase vary from $U(0, \pi)$. Intuitively, this just corresponds to rotated sine curves as can be seen in the examples in Section 5 of the Appendix. (see most left column for best density estimate of the true density.)

In this experiment we are given a variable number of context points during testing time ranging from $15, 30$ and $50$. This is done to investigate the ability of metaCDE to adapt even in small dataset situations.

MetaCDE/Neural Process/MetaNN are trained with $15, 30$ and $50$ context points on the tasks and with $80$ target points during training. At testing time, we simply pass the data through our model without having to retrain on the new unseen dataset. Note that we report again the p-values of the Wilcoxon signed one-sided test and we can see that as we decrease the context points, our methods is significantly outperforming the other methods. See Section 5 of Appendix for additional

Each of the non meta learning models DDE, KCEF, $\epsilon$-KDE, LSCDE are trained separately on each new dataset as they cannot share information between tasks.

\subsection{Model specifications}
For our MetaCDE we used a $3$-hidden layer Neural Network with $tanh$ activation functions and $Adam$ optimizer for all of our feature maps $\phi_x, \phi_y, b_{\theta}$. We cross-validate on held out dataset, over 32 and 64 hidden nodes per layer and $\lambda = 1.0, 0.1, 0.01$ for the regularization parameter. We fixed the learning rate at $1\text{e}$-$3$. 
We also set $\kappa=10$ as suggested by \cite{gutmann2012noise}.
\begin{itemize}
	\item  KCEF: we used the CV function that was in built in their Github repository (optimizing the parameters from a range $[1e\text{-}1, 10]$ \url{https://github.com/MichaelArbel/KCEF}) as well as consulted the authors to make sure we are using their method correctly.
	\item  LSCDE: We CV for $\sigma$ in $\text{logspace}(-3, 5, 20)$ and $\lambda$ in $\text{logspace}(-5, 5, 20)$
	\item  $\epsilon$-KDE: We CV over $\epsilon$ in $\text{linspace}(0.1, 1, 15)$ and bandwidth in $\text{linspace}(0.01, 1, 15)$
	\item  DDE: We CV over the bandwidth of $0.5$ and $1.0$
\end{itemize}

\textbf{NOTE:} We also tried to use a standard RBF kernel to compute our CMEO and CME with 50 context points on the synthetic data, however, these results were not up to par with what we got with NN. We searched for lengthscales over a range 0.01 to 10, however, the results were not comparable on synth data (84.65 +-21.15), see Table in section 5.1 for comparison.

The reason for this is first of all, the stationarity of the Gaussian kernel and secondly, only having very little data to construct a good CMEO. Using NN and CMEs we were able to capture the density by training it across tasks and hence learning more efficient embeddings than a gaussian kernel could. In addition, deep kernels have recently shown impressive results \cite{liu2020learning} and hence using a NN as a feature map is well justified.

\section{Neural Network version of our method (MetaNN)}
Furthermore, in order to investigate the importance of the CMEO task representation, we developed a purely neural version of our proposed method, which we call MetaNN. It differs from MetaCDE solely in the task representation. While MetaCDE uses kernel embeddings formalism to represents the task using CMEO calculated on the context points, MetaNN uses the DeepSets \cite{zaheer2017deep} approach, where the context pairs $(x_i, y_i)$ are simply concatenated into a vector, and then passed through a neural network. The outputs are then averaged to obtain the task embedding to which any new $x_{target}$ is concatenated to obtain the ``neural'' equivalent to CME. This neural representation is then being pushed through a Feed forward network in order to have the same dimension as $\phi_y(y_{target})$. This ensures that we can take the inner product to compute $s_{\theta}$.

The same training procedure as MetaCDE follows. We note that the concatenation of $x$ and $y$ encodes the joint distribution, rather than the conditional as in MetaCDE. While such task representation does preserve the relevant information, it is susceptible to changes in the marginal of $x$ across tasks and conditional representations are intuitively better suited for the task of conditional density estimation. The experiments demonstrate the value of combining the CME formalism with neural representations and we obtain significantly better results with MetaCDE compared to MetaNN. In our experiments we simply used a 3 layer MLP and cross-validate over either 32, 64 hidden nodes and learning rates of $1\text{e-}3$ or $ 1\text{e-}4$.
\subsection{Comparison of MetaNN to MetaCDE}
Next we will compare the MetaNN and MetaCDE algorithms in order to investigate the importance of the conditional mean embedding operator. As mentioned above, we have now swapped out the computation of the CMEO for a NN for MeatNN. This representation is then concatenated with the new $x_{target}$ to give us an element in $\mathcal{H}_Y$. We test out MetaNN on the synthetic dataset described in the experimental section.

\begin{figure*}[!htp]
	\centering
	\includegraphics[width=0.6\textwidth]{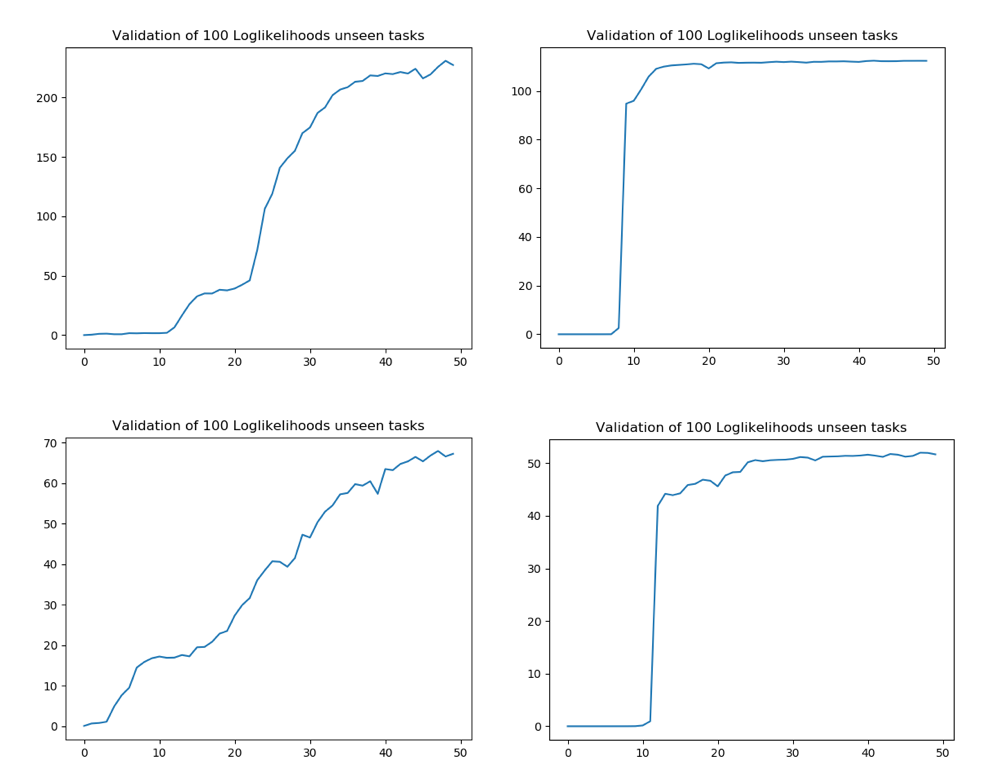}
	\caption{Figure illustrating how MetaNN performs better in low context points settings but seems to learn slower than MetaCDE. Top Row: 30 context points (left) MetaNN (right) MetaCDE; Bottom row: 15 context points (left) MetaNN (right) MetaCDE (x-axis represents 1 unit=10k tasks, y-axis loglikelihood)}
	\label{fig:like_app}
\end{figure*}

MetaNN will perform worse on $50$ context points as show in the main text but better on $15$, where MetaNN achieves a log-likelihood of $114.43 \pm 26.44$, whereas MetaCDE achieved only $ 51.73\pm 10.43$ (see Figure \ref{fig:like_app}). This can be explained by two factors. 

Firstly, a lower number of context points might give us a worse estimate of the conditional mean embedding operator. Secondly, we note that it takes significantly more task examples for the MetaNN to achieve the performance and hence this might have been due to the limited variety in the training task \textit{i.e.} variation in the range of the period and phase parameter. Hence we conjectured that MetaNN might have just memorized the tasks well. 

\begin{figure*}[!htp]
	\centering
	\includegraphics[width=0.6\textwidth]{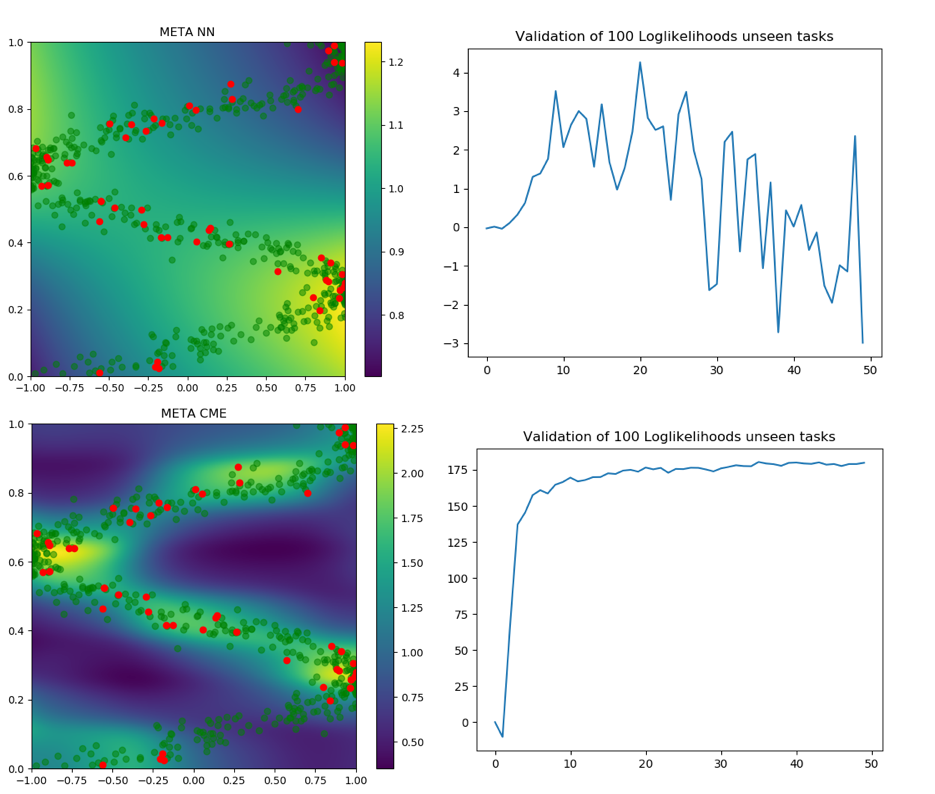}
	\caption{Figure illustrating how MetaNN fails when task become more variable. Top Row: MetaNN; Bottom row: MetaCDE (x-axis represents 1 unit=10k tasks, y-axis loglikelihood)}
	\label{fig:fail_cos}
\end{figure*}

Therefore we have ran additional experiments to on a harder synthetic dataset where we now sample $a$ from $U(4, 14)$ and phase vary from $U(-\pi, \pi)$ (i.e. more variety in the tasks than before where we sampled $a$ from $U(8, 12)$ and phase from $U(0, \pi)$). In this case, MetaNN seems to completely fail and not able to learn anything useful at all. As the tasks in this case are more variable (see figure \ref{fig:fail_cos}). Hence, we have not included MetaNN in the below figures when comparing with other conditional density methods.

\begin{figure*}[!htp]
	\centering
	\includegraphics[width=0.6\textwidth]{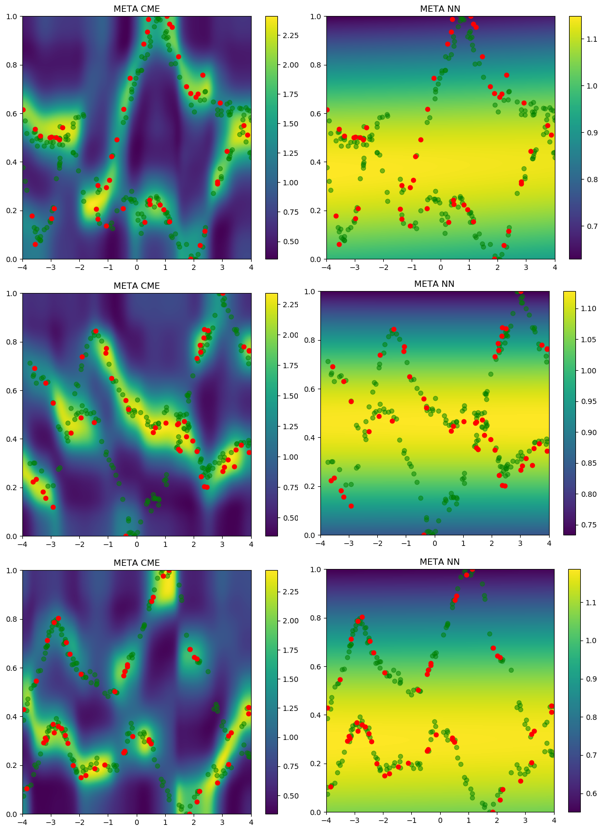}
	\caption{Density maps of the GP example (Left)MetaCDE (Right)(MetaNN)}
	\label{fig:gp_exp}
\end{figure*}

To further investigate this phenomenon, we have created a new task based on samples on Gaussian Processes (GPs). Here we sample 2 GPs with an Gaussian kernel with lengthscale 1 as well as a uniform random variable from $q \sim U(1, 3)$. We then added $u$ to one of the sampled GPs and hence created a multimodal dataset in $y$ (see figure \ref{fig:gp_exp}). This task has a lot more variability than the previous synthetic dataset task. Below, we illustrate how MetaCDE is still able to perform well whereas MetaNN completely fails to learn anything useful. This illustrates that CMEO includes useful additional inductive biases in our model. In particular, by using a CMEO, we explicitly tell the model which entries are covariates and which are responses and that the relevant property of the data for this task is the conditional distribution.

\begin{figure*}[!htp]
	\centering
	\includegraphics[width=0.6\textwidth]{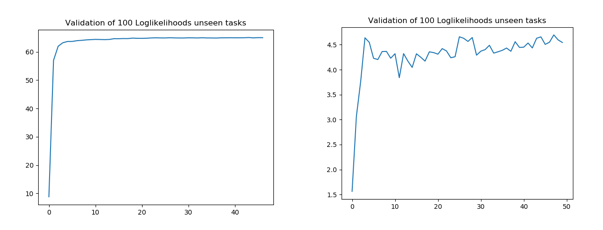}
	\caption{Evolution of the log-likelihood (x-axis represents 1 unit=10k tasks, y-axis loglikelihood) (Left)MetaCDE (Right)(MetaNN)}
	\label{fig:my_label}
\end{figure*}

\section{Normalization network $b_{\theta}$}
\begin{figure*}[!htp]
	\centering
	\includegraphics[width=0.5\textwidth]{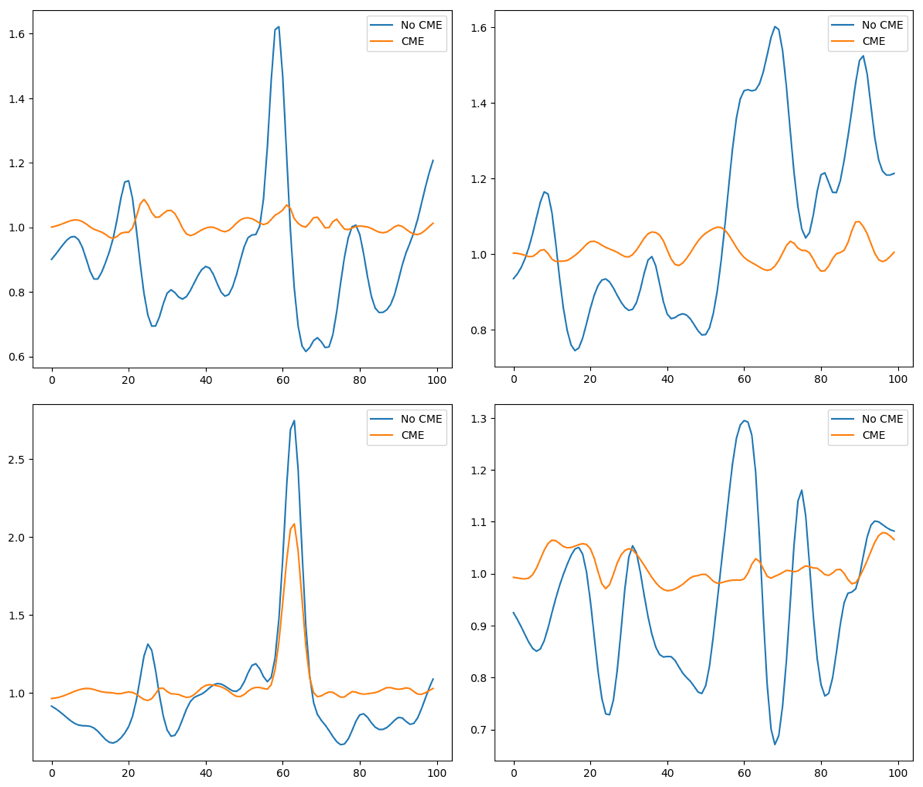}
	\caption{Post normalization needed after learning the density. Comparison shows the regimes where $b_{\theta}$ network includes CME as an input and the one where it does not.}
	\label{fig:norm_exp}
\end{figure*}

First of all, we want to note that learning the $b_{\theta}$ is not a novel idea but in fact has been proposed by the seminal paper on noise contrastive estimation \cite{gutmann2012noise}. \cite{gutmann2012noise} note that one can actually learn the normalization constant and show empirically that that the learnt density is actually close to a properly normalized density. Intuitively, the reason this happens is as follows. 
\begin{align}
	P_{\theta}(\text{True}|y, x) := \frac{p_\theta(y|x)}{p_\theta(y|x) + \kappa p_f(y)}.
	\label{classifier}
\end{align}
Assuming that we have a Bayes optimal classifier above and that $p_f$ is normalized (we choose this distribution so we know it is normalized), then assuming $p_\theta(y|x)$ was not normalized by a factor of $\gamma$ then we could just divide numerator and denominator by $\gamma$. This would hence just correspond to modifying $\kappa$ to $\kappa/\gamma$. However the Bayes optimal classifier sees exactly $\kappa$ more fake samples then real ones and hence in the case Bayes optimality was reached, we should have $\gamma$ close to 1, i.e. normalized $p_\theta(y|x)$. We even show in experiments that the normalziation needed is minimal (see Figure \ref{fig:norm_exp}). Additionally, we also noted that using $b_{\theta}$ actually helped in terms of stability in our training as it gives an extra degree of flexibility.

Next, we note that $b_\theta(x)$ depends on $s_{\theta}(x, y)$ and hence is task/context set dependent. Modelling $b_\theta(x)$ as a neural network that only takes input $x$ would not be task dependent, as it would be the same for each task and only depend on $x$. Therefore, we set the input of $b_\theta$ to be the CME $\widehat{\mu}_{Y=y|X=x}$, which thus incorporates all task information compactly (as $\widehat{\mu}_{Y=y|X=x}$ is computed using the CMEO). We have done further studies of this normalization network  and show how it effectively normalizes the density approximately.

To ensure a fair comparison between methods, we post-normalize the densities for all our experiments, i.e. whenever we compute a conditional density, $p(y^*|X=x)$, we first create a grid, $\{y_i\}_{i=1}^{100}$, and consider equally spaced evaluations $p(y_i|X=x)$ over the range of the data, and use them to re-normalize the density model before evaluating performance by computing the loglikelihood at $y^*$. 

We note that most methods are already producing density models that are very close to being normalized so the effect of the post-normalization is minimal. 

Fig. \ref{fig:norm_exp} illustrates the post-normalization in our GP experiments (i.e. the closer the line is to one the less we need to normalize our density), demonstrating that our approaches are able to learn an approximately normalized density. We however still perform the post normalization in order to remain fair compared to other methods. Another thing to note is that including the information of the CME $b_{\theta}(x)$ compared to not using it (no CME), allows us to have even less normalization necessary when learning the density. We have also found that the normalization network helps greatly in keeping the learned density approximately normalized.

\onecolumn
\section{Choice of the Fake Distribution in NCE}
The choice of the fake distribution plays a key role in the learning process here, especially because we are interested in conditional densities. In particular, if the fake density is significantly different from the marginal density $p(y)$, then our model could learn to distinguish between the fake and true samples of $y$ simply by constructing a ``good enough'' model of the marginal density $p(y)$ on a given task while completely ignoring the dependence on $x$ (this will then lead to feature maps that are constant in $x$). 

This becomes obvious if, say, the supports of the fake and the true marginal distribution are disjoint, where clearly no information about $x$ is needed to build a classifier -- i.e. the classification problem is ``too easy''. 

Thus, ideally we wish to draw fake samples from the true marginal $p(y)$ in a given task. While we could achieve this by drawing a $y$ paired to another $x$, i.e. from the empirical distribution of pooled $y$s in a given task, recall that we also require to be able to compute the fake density pointwise. Hence, we propose to use a kernel density estimate (KDE) of $y$'s as our fake density in any given task. 

In particular, a kernel density estimator of $p(y)$ is computed on all responses $y$ (context and target). To sample from the this fake density, we draw from the empirical distribution of pooled $y$'s and add Gaussian noise with standard deviation being the bandwidth of the KDE (we are using a Gaussian KDE for simplicity here; other choices of kernels are of course possible with appropriate modification of the type of noise). As our experiments demonstrate, this choice ensures that the fake samples are sufficiently hard to distinguish from the true ones, requiring the model to learn meaningful feature maps which capture the dependence between $x$ and $y$ and are informative for the CDE task.

Furthermore, we want to note that we chose $\kappa$ i.e. number of fakes samples to be $10$, mainly because in the original noise contrastive paper, they use also used $10$ and we noticed in our experiments that even when increasing $\kappa$ the performance didn't increase by much and hence just fixed $\kappa=10$.

Finally, we note that in principle it is possible to consider families of fake distributions which depend on the conditioning variable $x$. We do not explore this direction here and will leave this for future work. 

\newpage
\section{Illustration of Synthetic dataset}
In this section we illustrate that our proposed method, metaCDE, performs well even when data becomes scarce. Hence we applied our method on several tasks, where we have 50, 30 and 15 datapoints as context. In each of the experiments we also plotted the corresponding density, the method gave us and we can clearly see that only metaCDE is able to recover the "rotated sine curve" (synthetic data experiments) in all 3 cases, which is also shown in the loglikelihood estimates.
\subsection{Using 50 context points}
\begin{table*}[ht]
	\centering
	\resizebox{\columnwidth}{!}{%
		\begin{tabular}{lllllll}
			\hline
			& MetaCDE      & NP    & DDE                & LSCDE            & KCEF & $\epsilon$-KDE     \\ \hline
			\multicolumn{1}{l}{\textbf{Sythn. Data} Mean over 100 log-likelihoods} &  \textbf{197.84} $\pm$ \textbf{22.45} & -81.11$\pm$18.53  &162.98 $\pm$ 69.01 & 44.95 $\pm$ 74.36 & -388.30 $\pm$ 703.17   & 116.31 $\pm$ 236.99 \\ 
			\multicolumn{1}{l}{\textbf{Sythn. Data} P-value for Wilcoxon test}   & NA                  & < 2.2e-16  &8.144e-07             & <2.2e-16           & < 2.2e-16   &  2.384e-07   \\ \hline 
		\end{tabular}%
	}
\end{table*}
\begin{figure*}[ht]
	\centering
	\includegraphics[width=\textwidth]{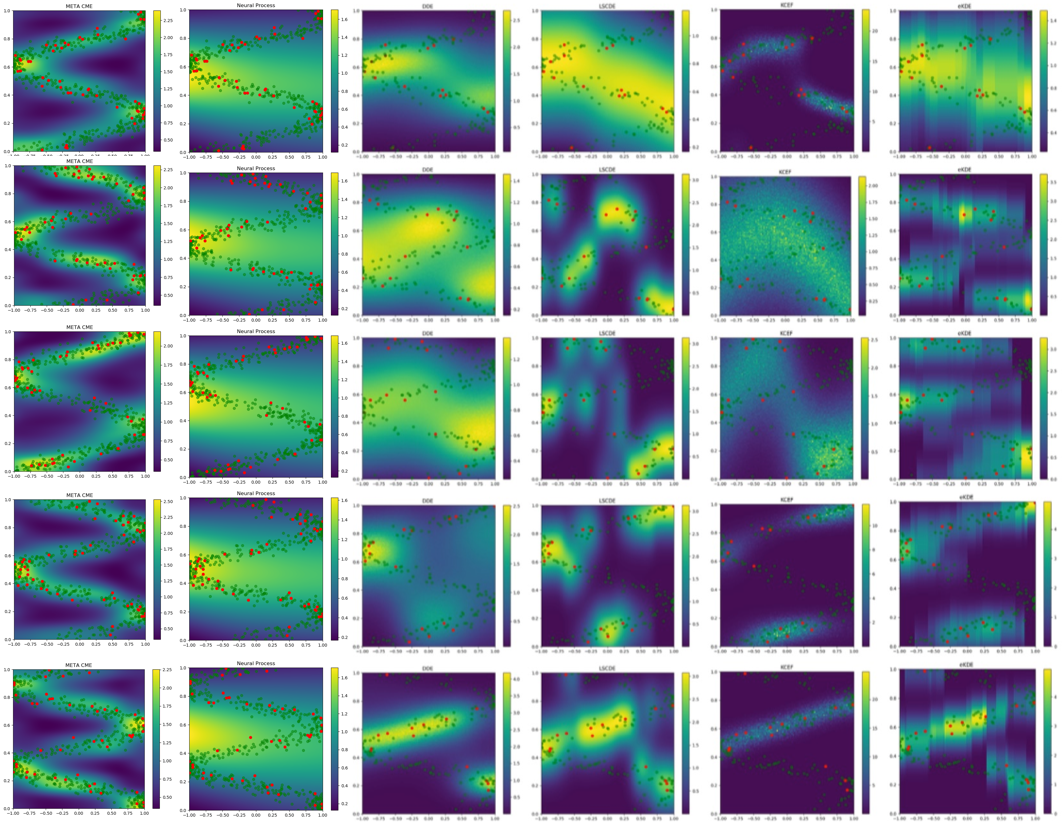}
	\caption{In Order (synthetic dataset): MetaCDE (ours), NP, DDE, LSCDE, KCEF, $\epsilon$-KDE \\
		The red dots are the context/training points and the green dots are points from the true density.}
\end{figure*}

\clearpage
\newpage
\onecolumn
\subsection{Using 30 context points}
\begin{table*}[ht]
	\centering
	\resizebox{\columnwidth}{!}{%
		\begin{tabular}{lllllll}
			\hline
			& MetaCDE      & NP    & DDE                & LSCDE            & KCEF & $\epsilon$-KDE     \\ \hline
			\multicolumn{1}{l}{\textbf{Sythn. Data} Mean over 100 log-likelihoods} &  
			\textbf{113.27} $\pm$ \textbf{17.36} & -48.98$\pm$12.26  &64.61 $\pm$ 54.33 & -23.02 $\pm$ 65.31  & -233.38 $\pm$ 528.99   & 29.64 $\pm$ 195.49 \\ 
			\multicolumn{1}{l}{\textbf{Sythn. Data} P-value for Wilcoxon test}   & NA                  & < 2.2e-16  & 4.577e-14            & <2.2e-16           & < 2.2e-16   &  4.917e-13   \\ \hline 
		\end{tabular}%
	}
\end{table*}

\begin{figure*}[ht]
	\centering
	\includegraphics[width=\textwidth]{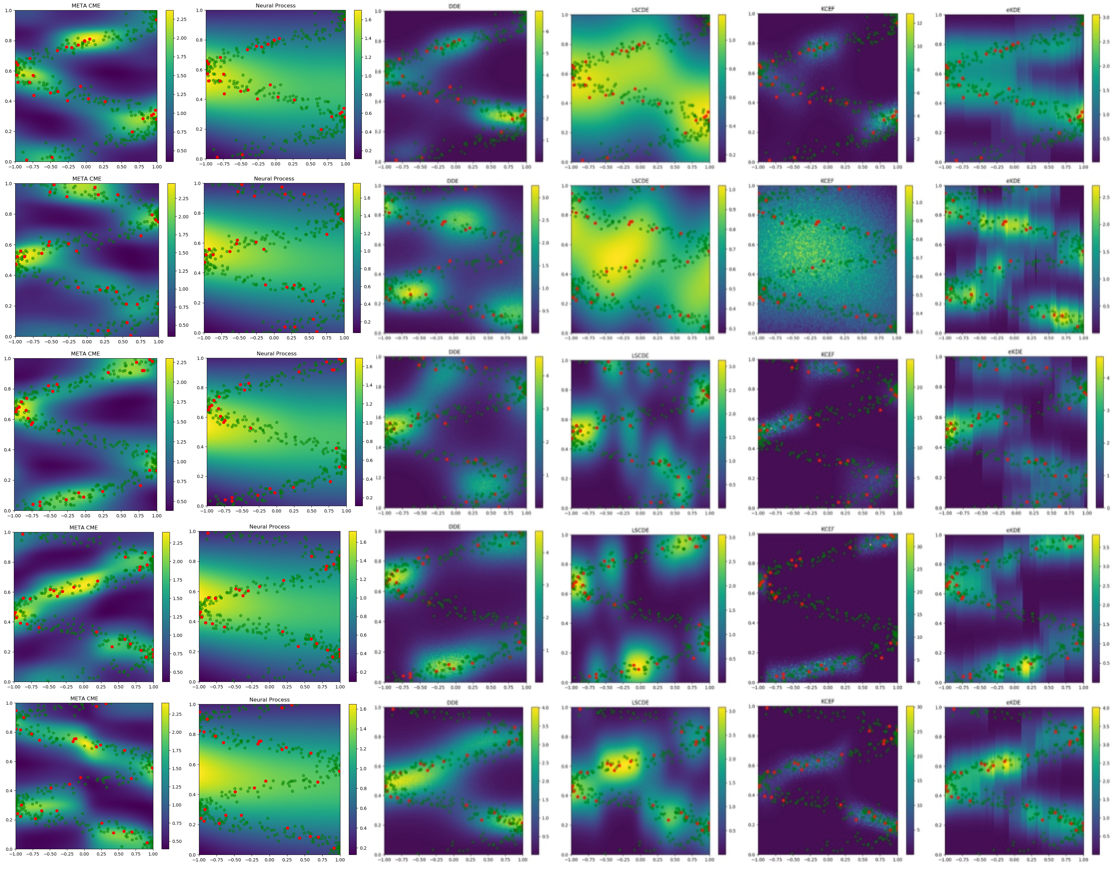}
	\caption{In Order (synthetic dataset): MetaCDE (ours), NP, DDE, LSCDE, KCEF, $\epsilon$-KDE \\
		The red dots are the context/training points and the green dots are points from the true density.}
\end{figure*}

\clearpage
\newpage
\onecolumn
\subsection{Using 15 context points}

\begin{table*}[ht]
	\centering
	\resizebox{\columnwidth}{!}{%
		\begin{tabular}{lllllll}
			\hline
			& MetaCDE      & NP    & DDE                & LSCDE            & KCEF & $\epsilon$-KDE     \\ \hline
			\multicolumn{1}{l}{\textbf{Sythn. Data} Mean over 100 log-likelihoods} &  
			\textbf{51.73} $\pm$ \textbf{10.48} & -24.39$\pm$8.20  & 0.58 $\pm$ 40.70 &-57.99 $\pm$ 59.13  & -142.19 $\pm$ 259.59  & -87.50 $\pm$ 224.13 \\ 
			\multicolumn{1}{l}{\textbf{Sythn. Data} P-value for Wilcoxon test}   & NA                  & < 2.2e-16  & 4.577e-14            & <2.2e-16           & < 2.2e-16   &  < 2.2e-16  \\ \hline 
		\end{tabular}%
	}
\end{table*}

\begin{figure*}[ht]
	\centering
	\includegraphics[width=\textwidth]{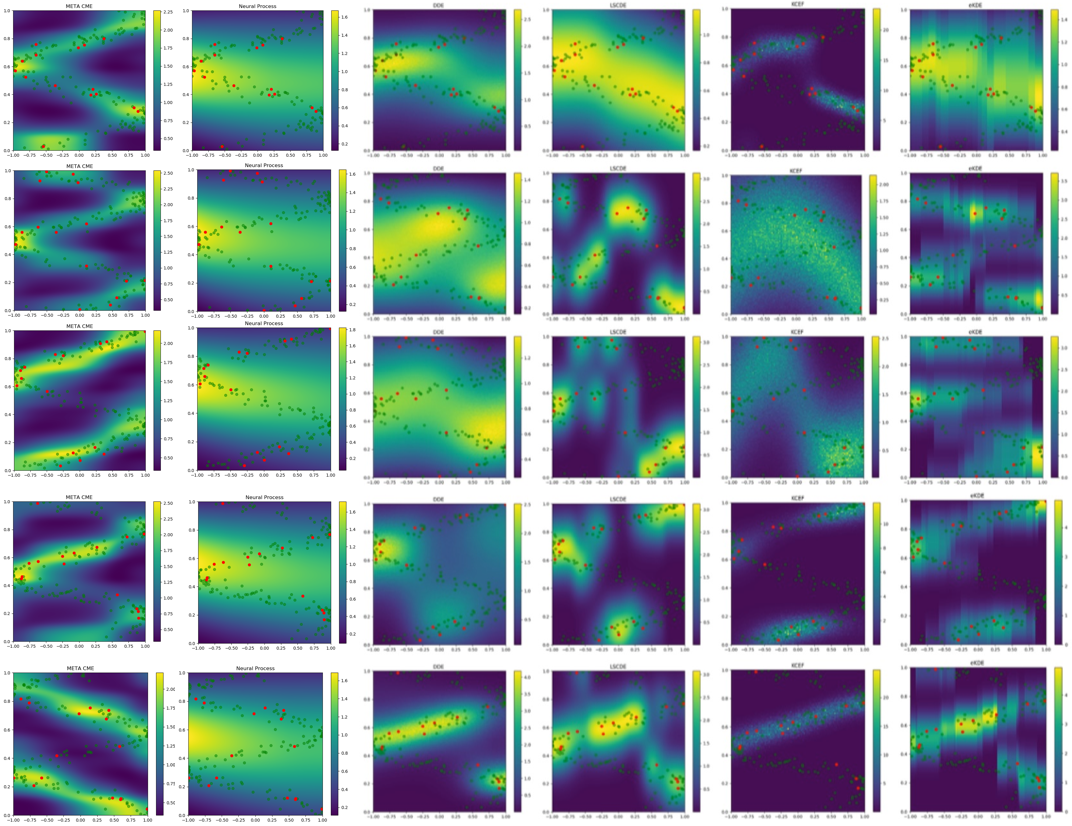}
	\caption{In Order (synthetic dataset): MetaCDE (ours), NP, DDE, LSCDE, KCEF, $\epsilon$-KDE \\
		The red dots are the context/training points and the green dots are points from the true density.}
\end{figure*}
\clearpage
\newpage
\onecolumn
\section{Further insight to the Ramachandran plots}
\subsection{Further details on Ramachandran plots}
To better understand the problem that we are tackling, consider the images below which represent different molecules with their respective fragments whose dihedral angle chemist measure. "\textit{A dihedral angle is the angle between two intersecting planes. In chemistry, it is the angle between planes through two sets of three atoms, having two atoms in common. - Wikipedia}".

\begin{figure*}[ht]
	\centering
	\includegraphics[width=0.5\textwidth]{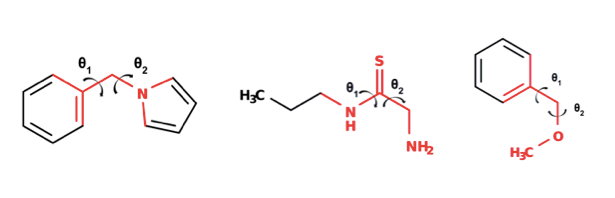}
	\caption{Example of molecules and the respective angles $\theta_1, \theta_2$ that we are trying to model. From this image it also becomes apparent how the multimodality arises as there are clearly some spatial symmetries. In the density plots below, the x-axis is $\theta_1$ and the y-axis is $\theta_2$}
\end{figure*}

\subsection{Additional information on the experimental setup}
In this experiment, we look into the Ramachandran plots for molecules.  Each plot indicates the energetically stable region of a pair of correlated dihedral angles in the molecule.  Specifically, we are interested in estimating the distributions of these correlated dihedral angles. We compute the conditional density for each correlated dihedral angles, given 20 context points at testing time. For our meta-learning training we use 20 context points and 60 targets points.

Note that the data was extracted from crystallography database \cite{gravzulis2011crystallography}. It is possible that some specific pairs of dihedral angles are rarely seen in the dataset, Hence, we may obtain a conditional density with high probability on the region without any observations in some cases. This is reasonable as the database covered only a small part of the chemical space and some potential area could be overlooked. Given that we assume that the support of our conditioning variable $x$ ranges from $[-\pi, \pi)$, we will inevitable also compute conditional distribution on areas where the configurations are not defined and hence the densities in those areas can be safely ignored as a computational chemist would not have queried these configurations in the first place.

\subsection{Model specifications}
For our MetaCDE we used a 3 hidden layer NN with $tanh$ activation functions for all of our feature maps. We cross validate over 32 and 64 hidden nodes per layer and $\lambda = 1.0, 0.1$  for the regularization parameter and over $1.0, 0.5, 0.3$ for the KDE lengthscale. We fix the learning rate at $1e$-$3$ and set $\kappa=10$.
\begin{itemize}
	\item  KCEF: we used the CV function that was in built in their Github repository (optimizing the parameters from a range $[1e-1, 10]$ \url{https://github.com/MichaelArbel/KCEF})
	\item  LSCDE: We CV for $\sigma$ in $\text{logspace}(-3, 5, 20)$ and $\lambda$ in $\text{logspace}(-5, 5, 20)$
	\item  $\epsilon$-KDE: We CV over $\epsilon$ in $\text{linspace}(0.5, 3, 15)$ and bandwidth in $\text{linspace}(0.01, 3, 15)$
	\item  DDE: We CV over bandwidth of $0.5$ and $1.0$
\end{itemize}
\textbf{NOTE:} Furthermore it looks like our method is not able to always capture the true trend given the limited amount of data. However, it seems to be able to capture some interesting \textbf{multimodal} patterns that would be useful to scientist to include in their models. Recently, there has been work done on these Ramachandran plots for Molecules by handcrafting the density maps. Our model allows us to compute the density maps without prior knowledge.

\begin{figure*}[ht]
	\centering
	\includegraphics[width=\textwidth]{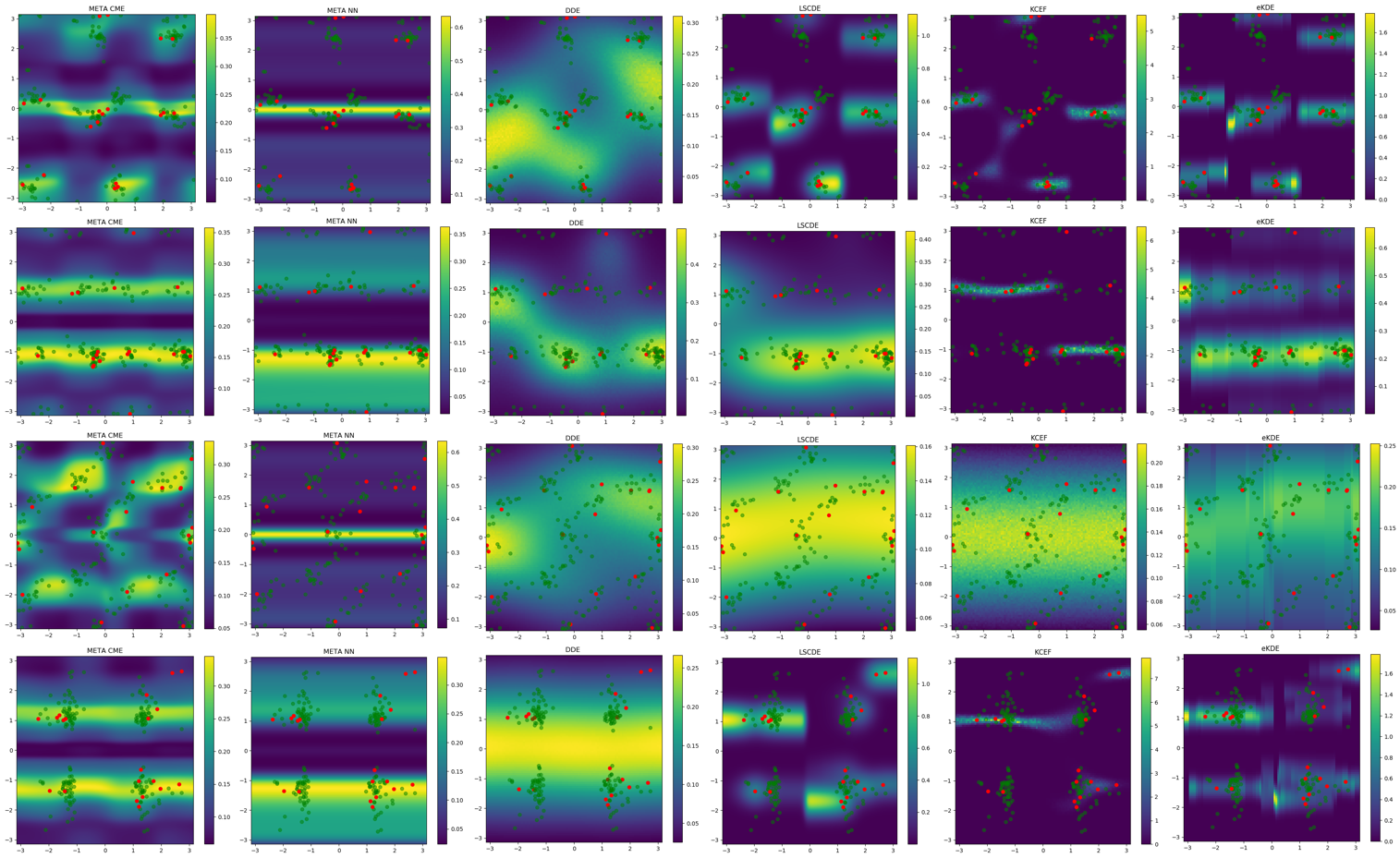}
	\caption{In Order (synthetic dataset): MetaCDE (ours), MetaNN (ours), NP, DDE, LSCDE, KCEF, $\epsilon$-KDE \\
		The red dots are the context/training points and the green dots are points from the true density.}
\end{figure*}

\clearpage
\newpage

\clearpage
\newpage
\onecolumn
\section{Illustration of the NYC taxi dataset}
\subsection{Experimental Setup}
We have extracted the publicly available dataset from the website \footnote{Data has been taken from: https://www1.nyc.gov/site/tlc/about/tlc-trip-record-data.page}.
We have first of all restricted ourselves to drop-off locations in from $-74.1$ to $-73.7$ in longitude and $40.6$ to $40.9$ in latitude. Next we have given our meta learning model 200 datapoints for context during training and 300 for target. At testing time we are presented with 200 context points and are required to compute the conditional density given a tip. In this case each task is one specific pickup location.
Again, we are using a $3$-hidden layer NN and CV over $32, 64, 128$ nodes and $\lambda = 0.1$ or $1.0$ and over $1.0, 0.5, 0.3$ for the KDE lengthscale. We use the $Adam$ optimizer and fixed the learning rate to $1\text{e}-3$. We also set $\kappa=10$.

\subsection{Note on the dataset}
In the main text we have seen how the drop-off density changes as we increase the amount of tips. This move of density illustrates well the data itself, as one is more likely to pay higher tips for longer journeys. Below we have plotted the drop-off locations of one specific pickup location colored with the respective tips paid.
\begin{figure*}[htp]
	\centering
	\includegraphics[width=\textwidth]{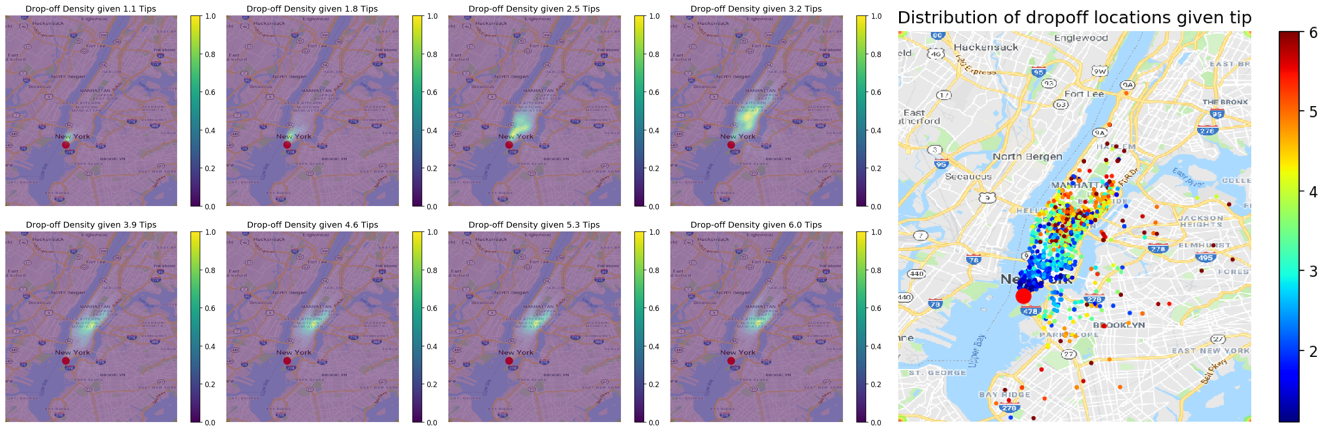}
	\caption{Density of the dropoff locations with increase in tips (right) the context and target points corresponding to the pickup locations (Big red dot the the pickup location)}
	\label{fig:my_label}
\end{figure*}
\clearpage
\newpage
\begin{figure*}[htp]
	\centering
	\includegraphics[width=\textwidth]{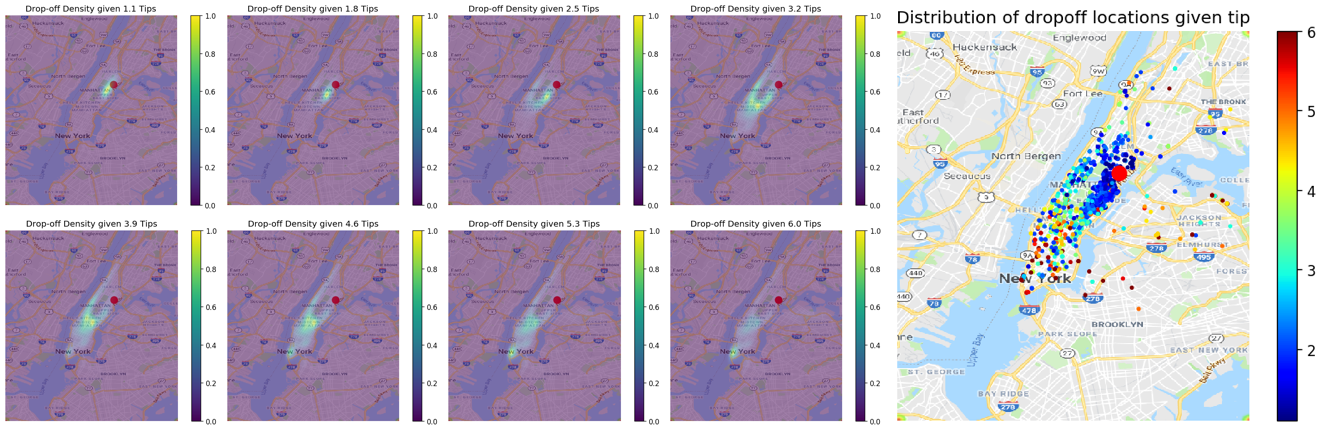}
	\caption{Density of the dropoff locations with increase in tips (right) the context and target points corresponding to the pickup locations (Big red dot the the pickup location)}
	\label{fig:my_label}
\end{figure*}

\begin{figure*}[htp]
	\centering
	\includegraphics[width=\textwidth]{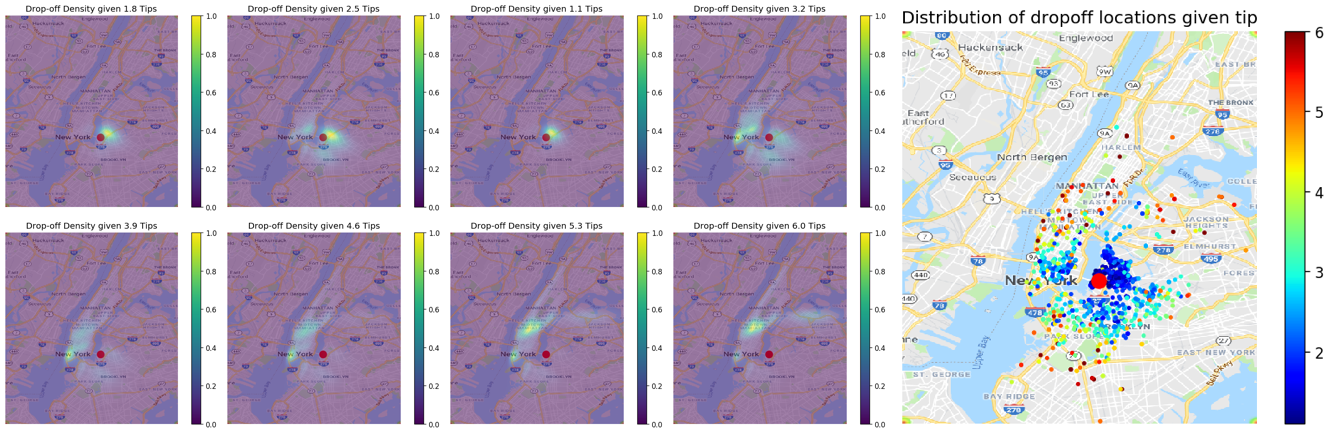}
	\caption{Density of the dropoff locations with increase in tips (right) the context and target points corresponding to the pickup locations (Big red dot the the pickup location)}
	\label{fig:my_label}
\end{figure*}

\end{document}